\documentclass{llncs}
\usepackage{amssymb}
\usepackage{amsmath}
\usepackage{latexsym}
\usepackage{graphicx}
\usepackage{multirow}
\usepackage{subfig}
\usepackage{wrapfig}
\usepackage[ruled, vlined]{algorithm2e}
\usepackage{float}
\usepackage{times}
\usepackage{url}

\newlength{\halftextwidth}
\numberwithin{equation}{section}

\floatstyle{ruled}
\newfloat{model}{htbp}{lop}

\def\disjunct#1{{b_{#1}}}
\def\job#1{{J_{#1}}}
\def\machine#1{{M_{#1}}}
\def\earliness#1{{E_{#1}}}
\def\lateness#1{{L_{#1}}}
\def\early#1{{e_{#1}}}
\def\late#1{{l_{#1}}}
\def\task#1{{t_{#1}}}
\def\dur#1{{p_{#1}}}
\def\lag#1{{L({#1})}}
\def\due#1{{d_{#1}}}
\def\we#1{{w^e_{#1}}}
\def\wt#1{{w^t_{#1}}}
\def\head#1{{h_{#1}}}

\def\mkp{C_{max}}
\def\etobj{ET_{sum}}
\def\tasks{{\cal T}}
\def\jobs{{\cal J}}
\def\machines{{\cal M}}

\title{Models and Strategies for Variants of the Job Shop Scheduling Problem}
\author{
Diarmuid Grimes\inst{1} \and
Emmanuel Hebrard\inst{2,3} 
}
\institute{
  Cork Constraint Computation Centre \\
  University College Cork, Ireland\\
  \texttt{d.grimes@4c.ucc.ie} 
  \and
  CNRS ; LAAS ; 7 avenue du colonel Roche, F-31077 Toulouse, France \\
  \and
  Universit\'e de Toulouse ; UPS, INSA, INP, ISAE ; UT1, UTM, LAAS ; F-31077 Toulouse, France \\
  \texttt{hebrard@laas.fr}
}

\begin{document}

\maketitle
\begin{abstract}


Recently, a variety of constraint programming and Boolean satisfiability 
approaches to scheduling problems have been introduced. 
They have in common the use of relatively simple
propagation mechanisms and an adaptive way to focus
on the most constrained part of the problem.
In some cases, these methods compare favorably to more classical
constraint programming methods relying on
propagation algorithms for global unary or cumulative resource constraints
and dedicated search heuristics.
 In particular,  we described an approach that
combines restarting, with a generic adaptive heuristic and solution guided branching
on a simple model based on a decomposition of disjunctive constraints.



In this paper, we introduce an adaptation of this technique for an
important subclass of job shop scheduling problems (JSPs), where the
objective function involves minimization of earliness/tardiness costs.
We further show that our technique can be improved by adding domain
specific information for one variant of the JSP (involving time lag
constraints). In particular we introduce a dedicated greedy heuristic,
and an improved model for the case where the maximal time
lag is 0 (also referred to as no-wait JSPs).

\end{abstract}

\section{Introduction}

Scheduling problems come in a wide variety and it is natural to think
that methods specifically engineered for each variant would
have the best performance. 
However, it was recently shown this is not always true.
Tamura et al. introduced an encoding of disjunctive and
precedence constraints into conjunctive normal 
form formulae \cite{tamura06:com}.
Thanks to this reformulation they
 were the first to report optimality proofs
for
all open shop scheduling instances from three widely studied
benchmarks. 
Similarly the hybrid CP/SAT solver \texttt{lazy-FD}~\cite{lazy-fd} was shown to be extremely
effective on Resource-Constrained Project scheduling (RCPSP) \cite{schutt09:why}.

Previously, we introduced an approach for open and job shop problems with a variety of extra constraints
\cite{grimes10:job,grimes09:clo}
using simple reified binary disjunctive constraints 
combined with a number of generic SAT and AI techniques: weighted degree variable ordering \cite{Boussemart-etal:2004b}, solution guided value ordering \cite{Becksgmp}, geometric restarting \cite{Walsh99} and nogood recording from restarts~\cite{LecoutreSTV07}.
It appears
that the weighted degree heuristic 
efficiently detects the most constrained parts of
the problem, 
focusing search on a fraction of the variables.

The simplicity of this approach makes it
easy to adapt to various constraints and objective functions.
One type of objective function that has proven troublesome for
traditional CP scheduling techniques involves minimizing the sum of
earliness/tardiness costs, primarily due to the weak propagation of
the sum objective \cite{Danna03lnsTechSchedetjsp}.
In this paper we show how our basic JSP model can be adapted to handle
this objective. Experimental results reveal that our approach is competitive with the state of the art on
the standard benchmarks from the literature.

Moreover, we introduce two refinements of our approach
for problems with maximum time lags between consecutive tasks, where we incorporate domain specific information to boost performance.
These time lag constraints, although conceptually
very simple, change the nature of the problem dramatically.
For instance, it is not
trivial to find a feasible schedule
even if we do not take into account any bound on the total makespan
(unless scheduling jobs back to back). 
This has several negative consequences. Firstly, 
it is not possible to obtain a trivial upper bound
of reasonable quality may be found by sequencing the
tasks in some arbitrary order.
The only obvious upper bound 
is to sequence the jobs consecutively. 
Secondly, since relaxing the makespan constraint
is not sufficient to make the problem easy, our 
approach can have difficulty finding a feasible 
solution for large makespans, even though it is very effective
when given a tighter upper bound. 
However because the initial upper bound is so poor,
even an exploration by dichotomy of the objective variable's domain
can take a long time.

We introduce a simple search strategy which,
when given a large enough upper bound on the makespan, 
 guarantees 
a limited amount of backtracking whilst still
providing good quality solutions. This simple strategy,
used as an initial step, greatly improves the performance
of our algorithm on this problem type. We report several 
new best upper bounds and proofs of optimality on
these benchmarks.
Moreover, we introduce another improvement in the
model of the particular case of \emph{No wait} JSP
where the tasks of each job must be directly consecutive.
This variant has been widely studied, and efficient 
metaheuristics have been proposed recently. 
 We report $5$
new best upper bound, and close $9$ new instances
in standard data sets.

Finally, because there are few comparison methods in the literature
for problems with strictly positive time lags, we adapted a job shop
scheduling model written in Ilog Scheduler by Chris Beck
\cite{Becksgmp}, to handle time lag constraints.
Our method outperforms this model when
time lag constraints are tight (short lags), however
when time lags are longer, the Ilog Scheduler model
together with geometric restarts and solution guided search
is better than our method.

\section{Background \& Previous work}
\label{sec:bac}

 An $n \times m$ job shop problem (JSP) involves 
a set of $nm$ \emph{tasks} $\tasks = \{\task{i} ~|~ 1 \leq i \leq nm\}$, partitioned into
$n$ \emph{jobs} $\jobs = \{\job{x} ~|~ 1 \leq x \leq n\}$, that need 
to be scheduled on $m$ \emph{machines} $\machines = \{\machine{y} ~|~ 1 \leq y \leq m\}$.
Each job $\job{x} \in \jobs$ is a set of $m$ tasks
$\job{x} = \{\task{(x-1)*m+y} ~|~ 1 \leq y \leq m\}$. 
Conversely, each machine $\machine{y} \in \machines$
denotes a set of $n$ tasks (to run on this machine) such that:
$\tasks = (\bigcup_{1 \leq x \leq n}\job{x}) = (\bigcup_{1 \leq y \leq m}\machine{y})$.

Each task $\task{i}$ has an associated duration, or processing time, $\dur{i}$.
A \emph{schedule} is a mapping of tasks to time points consistent with
sequencing and resource constraints. The former ensure that the tasks of each job
run in a predefined order whilst the latter
ensure that no two tasks
run simultaneously on any given machine.
In the rest of the paper, we shall identify
each task $\task{i}$ with the variable standing for its start time in the schedule.
We define the sequencing (2.1) and resource (2.2) constraints in Model~\ref{mod:jsp}.

Moreover, we shall consider two objective functions:
\emph{total makespan}, and \emph{weighted earliness/tardiness}.
In the former, we want to minimize the
 the total duration to run all tasks, 
that is, $\mkp = max_{\task{i} \in \tasks}(\task{i}+\dur{i})$
if we assume that we start at time $0$.
In the latter, each job $\job{x} \in \jobs$ has a due date, $\due{x}$. There
is a linear cost associated with completing a job before its due date,
 or the tardy completion of a job, 
with coefficient $\we{x}$ and $\wt{x}$, respectively. 
(Note that these problems differ from Just in Time job shop scheduling
problems\cite{Baptiste08SchedJIT}, where each \emph{task} has a due date.)
If $\task{xm}$ is the last task of job $\job{x}$, then
$\task{xm}+\dur{xm}$ is its completion time, hence
the cost of a job is then given by:
$
\etobj = \sum_{\job{x} \in \jobs}(max(\we{x} (\due{x}-\task{xm}-\dur{xm}), \wt{x}  (\task{xm}+\dur{xm}-\due{x})))
$

\begin{model}[htbp]
\footnotesize
\caption{\label{mod:jsp} JSP}
\vspace{-.15cm}
\begin{eqnarray}
  \task{i} + \dur{i} \leq \task{i+1} & \qquad \forall \job{x} \in \jobs,~ \forall \task{i}, \task{i+1} \in \job{x} \\
  \task{i} + \dur{i} \leq \task{j} ~\vee~ \task{j} + \dur{j} \leq \task{i} & \qquad \forall \machine{y} \in \machines,~ \task{i} \neq \task{j} \in \machine{y}  
\end{eqnarray}
\end{model}

\subsection{Boolean Model}
\label{ssec:model}

In previous work~\cite{grimes09:clo} we described the following simple model for open shop and job shop
scheduling.
First, to each task, we associate a variable $\task{i}$ taking its value in $[0, \infty]$ that stands for its
    starting time. 
Then, for each pair of tasks sharing a machine we introduce a Boolean variable
    that stands for the relative order of these two tasks.
    More formally, for each machine $\machine{y} \in \machines$,
    and for each pair of tasks $\task{i}, \task{j} \in \machine{y}$, we have 
    a Boolean variable $\disjunct{ij}$, and constraint (2.2) can be reformulated as follows:
    \begin{eqnarray}
    \disjunct{ij} = \left\{
      \begin{array}{cl}
        0 \Leftrightarrow \task{i} + \dur{i} \leq \task{j}\\
        1 \Leftrightarrow \task{j} + \dur{j} \leq \task{i}
      \end{array}\right. & \qquad \forall \machine{y} \in \machines,~ \task{i} \neq \task{j} \in \machine{y}  
    \end{eqnarray}
   Finally,  the tasks of each job $\job{x}$,
      are kept in sequence with a set of simple precedence constraints
      $\task{i} + \dur{i} \leq \task{i+1}$ for all $\task{i}, \task{i+1} \in \job{x}$.


For $n$ jobs and $m$ machines, this model therefore involves
 $nm(n-1)/2$ Boolean variables, as many disjunctive constraints, and $n(m-1)$ precedence constraints.
Bounds consistency (BC) is maintained on all constraints. Notice that state of the art
CP models use instead $m$ global constraints to reason about unary resources. 
The best known algorithms for filtering unary resources constraints
implement
 the edge finding, not-first/not-last, and detectable 
precedence rules with a $O(n \log n)$ time complexity \cite{Vilim2008Unary}. 
One might therefore expect our model 
to be less efficient as $n$ grows. However,
the quadratic number of 
constraints -- and Boolean variables -- required to model a resource 
in our approach has not proven
problematic on the academic benchmarks tested on to date.

\subsection{Search Strategy}

We refer the reader to \cite{grimes10:job} for a more detailed
description of the default search strategy used for job shop
variants, and we give here only a brief overview.

This model does not involve any global constraint associated to
a strong propagation algorithm. However, it appears that 
decomposing resource constraints into binary disjunctive 
elements is synergetic with adaptive heuristics, and 
in particular the
\emph{weighted-degree}-based heuristics \cite{Boussemart-etal:2004b}.
(We note that the greater the minimum arity of constraints
in a problem, the less discriminatory the weight-degree heuristic can be.) 
A constraint's weight is incremented by one each time
the constraint causes a failure during search. 
This weight can then be projected on variables to inform
the heuristic choices.

It is sufficient to decide the relative sequencing of the tasks,
that is, the value of the Boolean variables standing for disjuncts.
Because the domain size of these variables are all equal,
we use a slightly modified version of the \emph{domain over weighted-degree} heuristic,
where weights and domain size are taken on the two tasks whose relative ordering is
decided by the Boolean variable.
Let $w(\task{i})$ be the number of
times search failed while propagating any constraint
involving task $\task{i}$, and let 
$min(\task{i})$ and $max(\task{i})$ 
be, respectively, 
the minimum and maximum
 starting time of $\task{i}$ at any point during search. 
The next disjunct $\disjunct{ij}$ to branch on is the one
minimizing the value of:
\[
(max(\task{i}) + max(\task{j}) - min(\task{i}) - min(\task{j}) + 2) / (w(\task{i}) + w(\task{j}))
\] 

A second important aspect is the use of restarts. 
It has been observed that weighted heuristics also
have a good synergy with restarts \cite{grimes08:stu}.
Indeed, failures tend to happen at a given depth 
in the search tree, and therefore on constraints
that often do not involve variables corresponding to the first 
few choices. As a result, early restarts will tend 
to favor diversification until enough weight has been 
given to a small set of variables, on which the
search will then be focused.
We use a geometric restarting strategy \cite{Walsh99} with random tie-breaking. 
The geometric strategy is of the form $s, sr, sr^2, sr^3,...$ where $s$ is the base 
and $r$ is the multiplicative factor. In our experiments the base was 256 failures and the multiplicative factor was 1.3. 
Moreover, after each restart, the dead ends of the previous explorations are
stored as clausal nogoods~\cite{LecoutreSTV07}.

A third very important feature is the idea of 
guiding search (branching choices) based on the best solution found so far.
This idea is a simplified version of the solution 
guided approach (SGMPCS) proposed by Beck for JSPs \cite{Becksgmp}. 
Thus our search strategy can be viewed as variable ordering guided by past failures and value ordering guided by past successes.

Finally, before using a standard Branch \& Bound procedure,
we first use a dichotomic search to reduce the gap between lower and upper bound.
At each step of the dichotomic search, a
satisfaction problem is solved, with a limit on the number of nodes.

\section{Job Shop with Earliness/Tardiness Objective}
\label{sec:etjsp}

In industrial applications, the length of the makespan is not
always the preferred objective.
An  important alternative criterion 
 is the minimization of the cost of a job finishing
early/late. An example of a cost for early completion of a job would
be storage costs incurred, while for late completion of a job these
costs may represent the impact on customer satisfaction. 


Although the only change to the problem is the objective function, our
model requires a number of additional elements. 
When we minimize the sum of earliness and tardiness, we introduce
$4n$ additional variables. 
For each job $\job{x}$ we have a Boolean variable $\early{x}$ that
takes 
 the value $1$ iff $\job{x}$ is finished early and the value $0$ otherwise.
In other words,
$\early{x}$ is a reification of the precedence constraint $\task{xm} + \dur{xm} < \due{x}$.
Moreover, we also have a variable $\earliness{x}$ standing for the 
 duration between the completion time of the last task of $\job{x}$ and
the due date $\due{x}$ when $\job{x}$ is finished early:
$\earliness{x} = \early{x} (\due{x}-\task{xm}-\dur{xm})$.
Symmetrically, for each job $\job{x}$ we have Boolean variable $\late{x}$ taking
the value $1$ iff $\job{x}$ is finished late,
and an integer variable $\lateness{x}$ standing for the delay (Model~\ref{mod:etjsp}).


\begin{model}
\footnotesize
\caption{\label{mod:etjsp} ET-JSP}
\vspace{-.15cm}
\begin{eqnarray}
\nonumber   minimise~\etobj~~{\rm subject~to:} & \\
\etobj = \sum_{\job{x} \in \jobs}( \we{x} \earliness{x} + \wt{x} \lateness{x} ) \\
\early{x} \Leftrightarrow (\task{xm} + \dur{xm} < \due{x}) & \qquad \forall \job{x} \in \jobs \\
 \earliness{x} = \early{x} (\due{x}-\task{xm}-\dur{xm}) & \qquad \forall \job{x} \in \jobs\\
\late{x} \Leftrightarrow (\task{xm} + \dur{xm} > \due{x}) & \qquad \forall \job{x} \in \jobs \\
 \lateness{x} = \late{x} (\task{xm}+\dur{xm}-\due{x}) & \qquad \forall \job{x} \in \jobs \\
\nonumber ({\rm constraints}~ 2.1) ~\&~ ({\rm constraints~} 2.3)
\end{eqnarray}
\end{model}

Unlike the case where the objective involves minimizing the makespan,
branching only on the disjuncts is not sufficient for these
problems. Thus we also branch on the early and late Boolean variables,
and on the variables standing for start times of the last tasks of each job. 
For these extra variables, we use the standard definition of
domain over weighted degree.


\section{Job Shop Scheduling Problem with Time Lags}
\label{sec:tljsp}

Time lag constraints arise in many scheduling applications. 
For instance, in the steel industry,
 the time lag between the heating of a piece of steel 
and its moulding should be small~\cite{wismer72:sol}.
Similarly when scheduling chemical reactions, 
the reactives often cannot be stored
for a long period of time between two stages 
of a process to avoid interactions with 
external elements~\cite{rajendran94:ano}.

\subsection{Model}

The objective to minimise is represented by a variable $\mkp$ 
linked to 
 the last task of each job by $n$ precedence constraints:
$\forall x \in [1,\ldots,n]~ \task{xm} + \dur{xm} \leq \mkp$.
The maximum time lag between two consecutive tasks is simply
modelled by a precedence constraint with negative offset.
 Letting $\lag{i}$ be the maximum time lag between 
the tasks $\task{i}$ and $\task{i+1}$, we use the following model:

\begin{model}[htbp]
\footnotesize
\caption{\label{mod:jtl} TL-JSP}
\vspace{-.15cm}
\begin{eqnarray}
\nonumber   minimise~\mkp~~{\rm subject~to:} & \\
\mkp \geq \task{xm} + \dur{xm} & \qquad \forall \job{x} \in \jobs \\
\task{i+1} - (\dur{i} + \lag{i}) \leq \task{i} & \qquad \forall \job{x} \in \jobs,~ \forall \task{i}, \task{i+1} \in \job{x} \\
\nonumber ({\rm constraints}~ 2.1) ~\&~ ({\rm constraints~} 2.3)
\end{eqnarray}
\end{model}

\subsection{Greedy Initialization}

In the classical job shop scheduling problem, one can 
consider tasks in any order compatible with the jobs
and schedule them
to their earliest possible start time. The resulting
schedule may have a long makespan, however such a procedure
usually produces reasonable upper bounds.
With time lag constraints, however, scheduling early tasks
of a job implicitly fixes the start times for later tasks, thus making the problem
harder. Indeed, as soon as tasks have been fixed in several jobs,
the problem becomes difficult even if there is no constraint on the 
length of the makepsan.
Standard heuristics can thus have difficulty finding feasible solutions even when the makespan is not tightly bounded.
In fact, we observed that this phenomenon is 
critical for our approach.

Once a relatively good upper bound is known
our approach is efficient 
and is often able to find an optimal solution. However,
when the upper bound is, for instance, the trivial
sum of durations of all tasks, finding a feasible solution with 
such a relaxed makespan 
was in some cases difficult.
For some large instances, no  non-trivial solution was found, 
and on some instances  of more moderate size, much computational effort
was spent converging towards optimal values.

We therefore designed a search heuristic to find
solutions of good quality, albeit very quickly.
The main idea is to move to a new job only when all tasks of the same machine 
are completely sequenced between previous jobs.
Another important factor is to make decisions based on the maximum
completion time of a job, whilst leaving enough freedom
within that job to potentially insert subsequent jobs
instead of moving them to the back of the already scheduled jobs.

\begin{algorithm}
  \caption{\label{greedy}Greedy initialization branching heuristic}

  $fixed\_jobs \leftarrow \emptyset$;~
  $jobs\_to\_schedule \leftarrow \jobs$\;
  
  \While{$jobs\_to\_schedule \neq \emptyset$}{
    pick and remove a random job $\job{y}$ in $jobs\_to\_schedule$;~
    $fixed\_jobs \leftarrow fixed\_jobs \cup \{\job{y}\}$\;
    $next\_decisions \leftarrow \{\disjunct{ij} ~|~ \job{x(i)}, \job{x(j)} \in fixed\_jobs\}$\;
    \While{$next\_decisions \neq \emptyset$} {
      \lnl{dchoice} pick and remove a random disjunct $\disjunct{ij}$ from $next\_decisions$\;
      \lIf{$\job{x(i)} = \job{y}$} {
        branch on $\task{i} + \dur{i} \leq \task{j}$
      }
      \lElse{
        branch on $\task{j} + \dur{j} \leq \task{i}$\;
      }
    }
    \lnl{tchoice}branch on $\task{xm} \leq min(\task{(x-1)m+1})+$ \texttt{stretched}$(\job{y})$\;
  }
\end{algorithm}

We give a pseudo-code for this strategy in Algorithm~\ref{greedy}.
The set $jobs\_to\_schedule$ stands
for the jobs for which sequencing is still open,
whilst $fixed\_jobs$ contains the currently 
processed job, as well as all the jobs that
are completely sequenced. On the first iteration
of the outer ``while'' loop, a job is chosen.
There is no disjunct
satisfying the condition in Line~\ref{dchoice},
so this job's completion time is fixed to
a value given by the \texttt{stretched} procedure (Line~\ref{tchoice}), that
is, the minimum possible starting time of its first task,
plus its total duration, plus the sum of the possible 
time lags.

On the second iteration and beyond, a new job is
selected.
We then branch on the sequencing decisions between
this new job
and the rest of the set $fixed\_jobs$ 
before moving to a new job.
We call $\job{x(i)}$ the job that contains task $\task{i}$, and
observe that for any unassigned Boolean variable $\disjunct{ij}$, either 
$\job{x(i)}$ or $\job{x(j)} \in fixed\_jobs$ 
must be the last chosen job $\job{y}$. 
The sequencing choice that sets a task of the new job \emph{before}
a task of previously explored jobs is preferred, i.e., 
considered in the left branch.
Observe that a failure due to time lag constraints
can be raised only in the inner ``while'' loop. Therefore,
if the current upper bound on the makespan is large enough,
this heuristic will ensure that we never backtrack on 
a decision on a task.
We randomize this heuristic and use several iterations (1000 
in the present set of experiments) to find a good initial
solution.

\subsection{Special Case: Job Shop with no-wait problems}

The job shop problem with no-wait refers to the case where the maximum
time-lag is set to 0, i.e. each task of a job must start directly
after its preceding task has finished. In this case one can view the
tasks of the job as one block.

In \cite{grimes10:job} we introduced a simple improvement
for the no-wait class based on the following observation:
if no delay is allowed between any two consecutive tasks of a job,
then the start time of every task is functionally dependent
on the start time of any other task in the job.
The tasks of each job can thus be viewed as one block.
We therefore use a single variable $\job{x}$ standing
for the starting times of the job of same name. 

We call $\job{x(i)}$ the job of task $\task{i}$, and we define 
 $\head{i}$ as the total duration of the tasks coming before 
task $\task{i}$ in its job $\job{x(i)}$.
That is, $\head{i} = \sum_{k \in \{k ~|~ k < i ~\wedge~ \task{k} \in \job{x(i)}\}}\dur{k}$.
For every pair of tasks $\task{i} \in \job{x}, \task{j} \in \job{y}$ sharing a machine,
we use the same Boolean variables to represent disjuncts as in the 
original model, however linked by the following constraints:
\[
\disjunct{ij} = \left\{
  \begin{array}{cl}
    0 \Leftrightarrow \job{x} + \head{i} + \dur{i} - \head{j} \leq \job{y} \\
    1 \Leftrightarrow \job{y} + \head{j} + \dur{j} - \head{i} \leq \job{x} 
  \end{array}\right.
\]

Although the variables and constants are different, these
are the same constraints as used in the basic model. 
The no-wait JSP can therefore be reformulated
as shown in Model~\ref{mod:now1}, where the variables $\job{1}, \ldots, \job{n}$ represent
the start time of the jobs and $f(i,j) = \head{i} + \dur{i} - \head{j}$.

\begin{model}
\footnotesize
\caption{\label{mod:now1} NW-JSP}
\vspace{-.15cm}
\begin{eqnarray}
\nonumber   minimise~\mkp~~{\rm subject~to:} & \\
\mkp \geq \job{x} + \sum_{\task{i} \in \job{x}}\dur{i} & \qquad \forall \job{x} \in \jobs  \\
\disjunct{ij} = \left\{
\begin{array}{cl}
  0 \Leftrightarrow \job{x(i)} + f(i,j) \leq \job{x(j)}\\
  1 \Leftrightarrow \job{x(j)} + f(j,i) \leq \job{x(i)}
\end{array}\right. & \qquad \forall \machine{y} \in \machines,~ \task{i} \neq \task{j} \in \machine{y}
\end{eqnarray}
\end{model}

However, we can go one step further. For a given pair of jobs $\job{x},\job{y}$
the set of disjunct between tasks of these jobs define as many 
\emph{conflict intervals}
for the start time of one job relative to the other.
For two tasks $\task{i}$ and $\task{j}$, we have
$\job{x(j)} \not\in~]\job{x(i)} - f(j,i), \job{x(i)} + f(i,j)[$.
However, these intervals may overlap or subsume each other. 
It is therefore possible to tighten this encoding by computing
larger intervals, that we shall refer to as \emph{maximal forbidden intervals},
hence resulting in fewer
disjuncts.
We first give an example, and then briefly describe a procedure 
to find maximal forbidden intervals.

\begin{figure}[h]
  \centering
  \subfloat[\label{fig:fintervals:a}Sample problem.]{
    \includegraphics[scale=0.26]{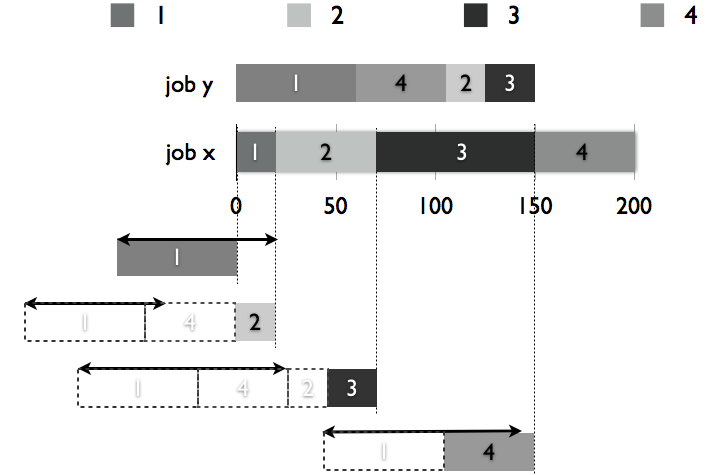} 
  }
  \hfill
  \subfloat[\label{fig:fintervals:b}Values of $\dur{}$, $\head{}$ and $f{}$.]{
    \begin{tabular}[b]{lcccc}
      ~&~ ~&~  ~&~  ~&~  \\
      Machine ~&~ $1$ ~&~ $2$ ~&~ $3$ ~&~ $4$ \\
      $\task{i},\task{j}$ ~&~ $\task{1},\task{5}$ ~&~ $\task{2},\task{7}$ ~&~ $\task{3},\task{8}$ ~&~ $\task{4},\task{6}$ \\
        $\dur{i}$ ~&~ 20 ~&~ 50 ~&~ 80 ~&~ 50 \\
        $\head{i}$ ~&~ 0 ~&~ 20 ~&~ 70 ~&~ 150 \\
        $\dur{j}$ ~&~ 60 ~&~ 20 ~&~  25 ~&~ 45 \\
        $\head{j}$ ~&~ 0 ~&~ 105 ~&~ 125 ~&~ 60 \\
        $-f(j,i)$ ~&~ -60 ~&~ -105 ~&~ -80 ~&~ 45 \\
      $f(i,j)$ ~&~ 20 ~&~ -35 ~&~ 25 ~&~ 140 \\
      ~&~ ~&~ ~&~ ~&~ \\
      ~&~ ~&~ ~&~ ~&~ \\
    \end{tabular}
  }
 \caption{\label{fig:fintervals} Computation of conflict intervals.}
\end{figure}

In Figure~\ref{fig:fintervals:a} we illustrate two jobs $\job{x} = \{\task{1}, \task{2}, \task{3}, \task{4}\}$ 
and $\job{y} = \{\task{5}, \task{6}, \task{7}, \task{8}\}$.
The number and shades of grey stand for the machine required
by each task. The length of the tasks are respectively $\{20,50,80,50\}$ for $\job{x}$
and $\{60,45,20,25\}$ for $\job{y}$. In Figure~\ref{fig:fintervals:b} we
give, for each machine, the pair of conflicting tasks, their durations
and the corresponding forbidden intervals.

For each machine $\machine{k}$, let $\task{i}$ be the task of $\job{x}$ and 
$\task{j}$ the task of $\job{y}$ that are both processed on machine $\machine{k}$.
Following the reasoning used in Model~\ref{mod:now1},
we have a conflict interval (represented by black arrows in Figure~\ref{fig:fintervals:a})
for each pair of tasks sharing the same machine:
$
\job{y} \not\in~ ]\job{x}-f(j,i), \job{x}+f(i,j)[
$.
In the example 
the forbidden intervals for $\job{y}$ are therefore:
$]\job{x}-60, \job{x}+20[ \ldots ]\job{x}-105, \job{x}-35[ \ldots ]\job{x}-80, \job{x}+25[ \ldots ]\job{x}+45, \job{x}+140[$. 
However, these intervals can be merged, yielding larger (maximal) forbidden intervals,
in which case we have:
$
\job{y} \not\in ]\job{x}-105, \job{x}+25[  ~\wedge~ \job{y} \not\in ]\job{x}+45, \job{x}+140[
$.

\begin{algorithm}
  \caption{\label{fint}\texttt{get-F-intervals}.}
  \KwData{$\job{x} = \{\task{x_1},\ldots,\task{x_m}\},~\job{y} = \{\task{y_1},\ldots,\task{y_m}\}, \machines$}
  $I_{in} \leftarrow []$\;
  \ForEach{$\task{x_i} \in \job{x}, \task{y_j} \in \job{y}$ such that $\machines(\task{x_i}) = \machines(\task{y_j})$} {
    $I_{in} \leftarrow I_{in}$~extended with~$[(-f(j,i), +1), (f(i,j), -1)]$\;
  }   
  \texttt{sort} $I_{in}$ by increasing first element\;
  $I_{out} \leftarrow []$;~
  $open \leftarrow 0$\;
  \While{\texttt{not-empty}$(I)$} {
    $(a,z) \leftarrow$~remove first element from $I_{in}$\;
    \lIf{$open = 0$}{append~$a$~to~$I_{out}$\;}
    $open \leftarrow open+z$\;
    \lIf{$open = 0$}{append~$a$~to~$I_{out}$\;}
  }
  return $I_{out}$\;
\end{algorithm}
Given two jobs $\job{x}$ and $\job{y}$, Algorithm~\ref{fint} computes all maximal forbidden intervals efficiently (in $O(m \log m)$ steps).
First, we build a list of pairs whose first element is an end point of a conflict interval, and second element is either $+1$ if it
is the start, and $-1$ otherwise.
Then these pairs are sorted by increasing first element. Now we can scan these pairs and count, thanks to the second
element, how many intervals are simultaneously open. When we go from $0$ to $1$ open intervals, this marks the start
of a maximal forbidden interval, and conversely the end when we go from $1$ to $0$ open intervals.
The list $I_{out}$ has $2k$ elements, and the $2i+1^{th}$ and $2i+2^{th}$ elements are read as the start and end of a forbidden interval.


Given this set of forbidden intervals, 
we can represent the conflicts between $\job{x}$ 
and $\job{y}$ with the
following set of Boolean variables and disjunctive constraints:

\begin{minipage}[b]{0.4\linewidth}
\centering
\[
\disjunct{xy}^{105,25} = \left\{
  \begin{array}{cl}
    0 \Leftrightarrow \job{y} + 105 \leq \job{x}\\
    1 \Leftrightarrow \job{x} + 25 \leq \job{y}
  \end{array}\right.
\]
\end{minipage}
\begin{minipage}[b]{0.4\linewidth}
\centering
\[
\disjunct{xy}^{45,140} = \left\{
  \begin{array}{cl}
    0 \Leftrightarrow \job{y} - 45 \leq \job{x}\\
    1 \Leftrightarrow \job{x} + 140 \leq \job{y}
  \end{array}\right.
\]
\end{minipage}

\medskip

In the previous encoding we would have needed $4$ Boolean variables and
as many disjunctive constraints (one for each pair of tasks sharing a machine).
We believe, however, that the main benefit is not the reduction in
size of the encoding. Rather, it is the tighter correlation between
the model and the real structure of the problem which helps
the heuristic to make good choices.

\begin{model}
\footnotesize
\caption{\label{mod:now2} NW-JSP}
\vspace{-.15cm}
\begin{eqnarray}
\nonumber   minimise~\mkp~~{\rm subject~to:} & \\
\mkp \geq \job{x} + \sum_{\task{i} \in \job{x}}\dur{i} & \qquad \forall \job{x} \in \jobs  \\
\disjunct{ij}^{a,b} = \left\{
\begin{array}{cl}
  0 \Leftrightarrow \job{y} -a \leq \job{x} \\
  1 \Leftrightarrow \job{x} +b \leq \job{y}
\end{array}\right. & \qquad \forall \job{x} \neq \job{y} \in \jobs, [a,b] \in \texttt{get-F-intervals}(\job{x}, \job{y}, \machines)
\end{eqnarray}
\end{model}



\section{Experimental Evaluation}
\label{sec:exp}

The full experimental results, with statistics for each instance, 
as well as benchmarks and source
code are online: \url{http://homepages.laas.fr/ehebrard/jsp-experiment.html}.

\subsection{Job Shop with Earliness/Tardiness Objective}
\label{exp:etjsp}


The best complete methods for handling these types of problem are the
CP/LP hybrid of Beck and Refalo \cite{Beck03Schedetjsp} and the MIP
approaches of Danna et
al. \cite{Danna03rinsSchedetjsp}, and Danna and Perron \cite{Danna03lnsTechSchedetjsp}, while
more recently Kebel and Hanzalek proposed a pure CP approach
\cite{Kelbel10etjspSched}. Danna and Perron also proposed an incomplete approach
based on large neighborhood search \cite{Danna03lnsTechSchedetjsp}.

Our experiments were run on an Intel Xeon 2.66GHz machine with
12GB of ram on Fedora 9. Each algorithm run on a problem had an
overall time limit of 3600s, and there were 10 runs per instance. We
report our results in terms of the best and worst run. We
tested our method on two benchmarks which have been widely studied in
the literature.
The comparison experimental results are taken from
\cite{Danna03rinsSchedetjsp} and \cite{Danna03lnsTechSchedetjsp}, where
all experiments were performed on a 1.5 GHz Pentium IV system running
Linux. For the first benchmark, these algorithms had a time limit of
20 minutes per instance, while for the second benchmark the algorithms
had a time limit of 2 hours.

The comparison methods are as follows:
\begin{itemize}

\item 
\emph{MIP}: Default CPLEX in \cite{Danna03rinsSchedetjsp}, run using a modified version of ILOG CPLEX 8.1 

\item 
\emph{CP}: A pure constraint programming approach introduced by Beck and Refalo in \cite{Beck03Schedetjsp}, run using ILOG Scheduler 5.3 and ILOG Solver 5.3

\item 
\emph{CRS-ALL}: A CP/LP hybrid approach proposed by Beck and Refalo in \cite{Beck03Schedetjsp}, run using ILOG CPLEX 8.1, ILOG Hybrid 1.3.1, ILOG Scheduler 5.3 and ILOG Solver 5.3

\item 
\emph{uLNS}: An unstructured large neighborhood search MIP method proposed by Danna and Peron in \cite{Danna03lnsTechSchedetjsp}, run using a modified version of ILOG CPLEX 8.1

\item 
\emph{sLNS}: A structured large neighborhood search CP/LP method proposed by Danna and Peron in \cite{Danna03lnsTechSchedetjsp}, run using ILOG Scheduler 5.3, ILOG Solver 5.3 and ILOG CPLEX 8.1
\end{itemize}



The first benchmark consists of 9 sets of problems 
generated by Beck and Refalo \cite{Beck03Schedetjsp} using the random
JSP generator of Watson et al. \cite{Watson99JspGeneratorSched}. For
instance size $\jobs$x$\machines$, there were three sets of ten JSPs of
size 10x10, 15x10 and 20x10 generated. 
The second benchmark is taken from the genetic algorithms (GA)
literature and was proposed by Morton and Pentico
\cite{Morton93DynJspSched}. There are 12 instances, with problem size
ranging from 10x3 to 50x8. Jobs in these problems do have release
dates. Furthermore earliness and tardiness costs of a job are equal.


We present results on the randomly generated ETJSPs in
Table~\ref{tab:ETJSPRandOpt} in terms of number of problems solved to
optimality and sum of the upper bounds, for each algorithm.\footnote{Note that sLNS is not complete, hence it never proved optimality.}
%
Here, the column ``Best'' for
our method means the number of problems solved to optimality on at
least one of the ten runs on the instance, while the column ``Worst'' refers to the
number of problems solved to optimality on all ten runs. We also report the mean
cpu time in seconds for our method.

\begin{table}[htbp]
  \footnotesize
  \caption{ET-JSP - Random Problems, Number Proven Optimal and Upper Bound Sum \label{tab:ETJSPRandOpt}}
  \begin{center}
    \begin{tabular}{|l|| r r | r r | r r | r | r r | | r r | r r | r |  }
\hline \hline
\multirow{2}{*}{$lf$} 
& \multicolumn{ 2 }{c|}{ \multirow{2}{*}{MIP} } 
& \multicolumn{ 2 }{c|}{ \multirow{2}{*}{CP} } 
& \multicolumn{ 2 }{c|}{ \multirow{2}{*}{uLNS} } 
& \multicolumn{ 1 }{c|}{ \multirow{2}{*}{sLNS} } 
& \multicolumn{ 2 }{c||}{ \multirow{2}{*}{CRS-All} } 
& \multicolumn{5}{c|}{Model~\ref{mod:etjsp}} \\ 
&  \multicolumn{2}{c|}{} 
&  \multicolumn{2}{c|}{} 
&  \multicolumn{2}{c|}{} 
&  \multicolumn{1}{c|}{} 
&  \multicolumn{2}{c||}{}  
&  \multicolumn{2}{c|}{Best} 
&  \multicolumn{2}{c|}{Worst}
&  \multicolumn{1}{c|}{Avg.} \\ \hline
& \multicolumn{1}{c}{opt.} & \multicolumn{1}{c|}{$\sum$ub} 
& \multicolumn{1}{c}{opt.} & \multicolumn{1}{c|}{$\sum$ub}
& \multicolumn{1}{c}{opt.} & \multicolumn{1}{c|}{$\sum$ub}
& \multicolumn{1}{c|}{$\sum$ub} 
& \multicolumn{1}{c}{opt.} & \multicolumn{1}{c||}{$\sum$ub}
& \multicolumn{1}{c}{opt.} & \multicolumn{1}{c|}{$\sum$ub} 
& \multicolumn{1}{c}{opt.} & \multicolumn{1}{c|}{$\sum$ub} 
& \multicolumn{1}{c|}{Time (s)} \\
      1.0 ~& 0 ~& 654,290 ~& 0 ~& 1,060,634 ~& 0 ~& 156,001 ~& 52,307 ~& 7 ~& 885,546 ~& \textbf{10} ~& \textbf{30,735} ~& 8  ~& 38,416 ~& 2534.86 \\
      1.3 ~& 14 ~& 26,930 ~& 6 ~& 1,248,618 ~& \textbf{30} ~& \textbf{8,397} ~& \textbf{8,397} ~& \textbf{30} ~& \textbf{8,397} ~& \textbf{30} ~& \textbf{8,397} ~& \textbf{30} ~& \textbf{8,397} ~& 0.36   \\
      1.5 ~& 27 ~& 7,891 ~& 6 ~& 1,672,511 ~& \textbf{30} ~& \textbf{6,964} ~& \textbf{6,964} ~& \textbf{30} ~& \textbf{6,964} ~& \textbf{30} ~& \textbf{6,964} ~& \textbf{30}  ~& \textbf{6,964} ~& 0.18 \\ \hline         \hline
\multicolumn{15}{l}{Notes: Comparison results taken from \cite{Danna03rinsSchedetjsp}, except uLNS, taken from \cite{Danna03lnsTechSchedetjsp}.}\\
\multicolumn{15}{l}{Figures in bold are the best result over all methods.}\\
    \end{tabular}
  \end{center}
\end{table}



We first consider the number of problems solved to optimality (columns ``Opt.''). 
While there is little difference in the performance of our method and
that of uLNS and CRS-ALL on the looser instances (looseness factor of
1.3 and 1.5), we see that our method is able to close three of the 23
open problems in the set with looseness factor 1.0. An obvious reason
for this improvement with our method would be the difference in time
limits and quality of machines. However, analysis of the results
reveals that of the 68 problems solved to optimality on every run of
our method, only 8 took longer than one second on average, and only one
took longer than one minute (averaging 156s). Furthermore, uLNS only
solved two problems to optimality when the time limit was increased to
two hours \cite{Danna03lnsTechSchedetjsp}. Clearly our method is
extremely efficient at proving optimality on these problems.

The previous results suggest that CRS-ALL is much better than uLNS on
these problems. However, as was shown by Danna et
al. \cite{Danna03rinsSchedetjsp}, this may not be the case when the
algorithms are compared based on the sum of the upper bounds found
over the 30 ``hard'' instances (i.e. with looseness factor 1.0). 
In order to assess whether there was a similar
deterioration in the performance of our method as for CRS-ALL on the
problems where optimality was not proven, we report this data
in the
columns ``$\sum$ub'' of
Table~\ref{tab:ETJSPRandOpt}.


 We find, on the contrary, that the performance of our approach is even
more impressive when algorithms are compared using this metric. The
two large neighborhood search methods found the best upper bounds of
the comparison algorithms with sLNS the most efficient by a factor of
2 over uLNS. However, there are a couple of points that should be
noted here. Firstly sLNS is an incomplete method so cannot prove
optimality, and secondly the sum of the worst upper bounds found by
our method was still significantly better than that found by sLNS.
Indeed, there was very little variation in performance for our method
across runs, with an average difference of 256 between the best and
worst upper bounds found.

Danna and Perron also provided the sum of the best upper bounds found
on the hard instances over all methods they studied
\cite{Danna03lnsTechSchedetjsp}, which was 36,459. This further underlines
the quality of the performance of our method on these problems. Finally,
we investigated the hypothesis that the different time limit and
machines used for experiments could explain these results. We compared
the upper bounds found by our method after the
dichotomic search phase, where the maximum runtime of this phase over
all runs per instance was 339s. The upper bound sums over the hard
instances were 32,299 and 49,808 for best and worst respectively,
which refutes this hypothesis.

\begin{table*}[htbp] 
  \footnotesize
  \caption{ET-JSP - GA Problems, Normalized upper bounds \label{tab:ETJSPGAobj}}
 \begin{center}
    \footnotesize
    \begin{tabular}{|lc||c||ccccc||cc|}
      \hline
      \hline
      \multicolumn{ 1 }{|l}{\multirow{2}{*}{Instance}} & \multicolumn{ 1 }{c||}{\multirow{2}{*}{Size}} & \multicolumn{ 1 }{c||}{ \multirow{2}{*}{MIP} } & \multicolumn{ 1 }{c}{ \multirow{2}{*}{CP} } & \multicolumn{ 1 }{c}{ \multirow{2}{*}{uLNS} } & \multicolumn{ 1 }{c}{ \multirow{2}{*}{sLNS} } & \multicolumn{ 1 }{c}{ \multirow{2}{*}{CRS-All} } & \multicolumn{ 1 }{c||}{ GA } & \multicolumn{2}{c|}{Model~\ref{mod:etjsp}} \\
      ~&~			~&~		~&~	~&~	~&~	~&~  ~&~	Best	~&~	Best	~&~ Worst	\\	\hline				
jb1	~&~	10x3		~&~	\textbf{0.191*}	~&~	0.474	~&~	\textbf{0.191*}	~&~	\textbf{0.191}	~&~	\textbf{0.191*}	~&~	0.474	~&~	\textbf{0.191*}	~&~	\textbf{0.191*}	\\
jb2	~&~	10x3		~&~	\textbf{0.137*}	~&~	0.746	~&~	\textbf{0.137*}	~&~	\textbf{0.137}	~&~	0.531	~&~	0.499	~&~	\textbf{0.137*}	~&~	\textbf{0.137*}	\\	
jb4	~&~	10x5		~&~	\textbf{0.568*}	~&~	0.570	~&~	\textbf{0.568*}	~&~	\textbf{0.568}	~&~	\textbf{0.568*}	~&~	0.619	~&~	\textbf{0.568*}	~&~	\textbf{0.568*}	\\	
jb9	~&~	15x3		~&~	\textbf{0.333*}	~&~	0.355	~&~	\textbf{0.333*}	~&~	\textbf{0.333}	~&~	1.216	~&~	0.369	~&~	\textbf{0.333*}	~&~	\textbf{0.333*}	\\	
jb11	~&~	15x5		~&~	0.233	~&~	0.365	~&~	\textbf{0.213*}	~&~	\textbf{0.213}	~&~	\textbf{0.213*}	~&~	0.262	~&~	0.221	~&~	0.235	\\	
jb12	~&~	15x5		~&~	\textbf{0.190*}	~&~	0.239	~&~	\textbf{0.190*}	~&~	\textbf{0.190}	~&~	\textbf{0.190*}	~&~	0.246	~&~	\textbf{0.190*}	~&~	\textbf{0.190*}	\\	\hline		~&~	GMR	~&~	1.015	~&~	1.774	~&~	\textbf{1}	~&~	\textbf{1}	~&~	1.555	~&~	1.610	~&~	1.006	~&~	1.017 \\ \hline\hline
ljb1	~&~	30x3		~&~	\textbf{0.215*}	~&~	0.847	~&~	\textbf{0.215*}	~&~	\textbf{0.215}	~&~	0.295	~&~	0.279	~&~	\textbf{0.215}	~&~	0.221	\\	
ljb2	~&~	30x3		~&~	0.622	~&~	1.268	~&~	\textbf{0.508}	~&~	\textbf{0.508}	~&~	1.364	~&~	0.598	~&~	0.590	~&~	0.728	\\	
ljb7	~&~	50x5		~&~	0.317	~&~	0.614	~&~	0.123	~&~	\textbf{0.110}	~&~	0.951	~&~	0.246	~&~	0.166	~&~	0.256	\\	
ljb9	~&~	50x5		~&~	1.373	~&~	1.737	~&~	1.270	~&~	1.015	~&~	2.571	~&~	\textbf{0.739}	~&~	1.157	~&~	1.513	\\	
ljb10	~&~	50x8		~&~	0.820	~&~	1.569	~&~	0.558	~&~	0.525	~&~	1.779	~&~	0.512	~&~	\textbf{0.499}	~&~	0.637	\\	
ljb12	~&~	50x8		~&~	1.025	~&~	1.368	~&~	0.488	~&~	0.605	~&~	1.601	~&~	\textbf{0.399}	~&~	0.537	~&~	0.623	\\	\hline
	~&~	GMR 	~&~	1.943	~&~	3.233	~&~	1.213	~&~	\textbf{1.170}	~&~	4.098	~&~	1.220	~&~	1.299	~&~	1.686		\\ \hline\hline
\multicolumn{2}{|c|}{Overall GMR}		&~	1.329	~&~	2.434	~&~	1.084	~&~	\textbf{1.068}	~&~	2.305	~&~	1.408	~&~	1.118	~&~	1.256	\\ \hline\hline
\multicolumn{10}{l}{Comparison results taken from \cite{Danna03rinsSchedetjsp}. Figures in bold indicate best upper bound}\\
\multicolumn{10}{l}{found over the different algorithms. ``*'' indicates optimality was proven by the algorithm.}\\
\end{tabular}
\end{center}
\end{table*}

Table~\ref{tab:ETJSPGAobj} provides results on the second of the
benchmarks (taken from the GA literature). Following the convention of
previous work on these problems
\cite{Vazquez00etjspSched}\cite{Beck03Schedetjsp}\cite{Danna03rinsSchedetjsp},
we report the cost normalized by the weighted sum of the job
processing times.  We include the best results found by the GA
algorithms as presented by V{\'a}zquez and Whitley
\cite{Vazquez00etjspSched}. We also provide an aggregated view of the results of each algorithm using the geometric mean ratio (GMR), 
which is the geometric mean of the ratio between the normalized upper bound found by the algorithm and the best known normalized upper bound, across a set of instances.

The performance of our method was less impressive for these problems,
solving two fewer problems to optimality than uLNS, and achieving a worse GMR than either of the large neighborhood search methods. 
However, we remind the reader that all comparison methods
had a 2 hour time limit on these instances, except the GA approaches
for which the time limit was not reported.
We further note that we find an improved solution for one instance (ljb10) and outperform all methods other than uLNS and sLNS.


\subsection{Job Shop Scheduling Problem with positive Time Lags}

These experiments were run using the same settings as in Section~\ref{exp:etjsp}.
However, because of the large number of instances and algorithms, we
used only 5 random runs per instance.


There are relatively few results reported for benchmarks
with positive maximum time lag constraints, as most publications 
focus on the ``no wait'' case.
Caumond et al. introduced a genetic algorithm~\cite{DBLP:journals/cor/CaumondLT08}.
Then, Artigues et al. introduced a Branch \& Bound procedure
that allowed them to find lower bounds of good quality~\cite{Artigues2011220}.
Therefore, in order 
to get a better idea of the efficiency of our approach, 
we adapted a model written by Chris Beck 
for Ilog Scheduler (version 6.3) 
 to problems featuring time lag constraints.
This model was used to showcase the SGMPCS algorithm \cite{Becksgmp}.
We used the following two strategies: 
In the first, the next pair of tasks to schedule is chosen following 
the Texture 
heuristic/goal predefined in Ilog Scheduler and restarts
following the Luby sequence~\cite{LubySZ93} are performed, 
this was one of the default
strategies used as a reference point in~\cite{Becksgmp}. In the second, 
branching decisions are selected with the same ``goal'', however the previous
best solution is used to decide wich branch should be explored first, and
geometric restarts~\cite{Walsh99} are performed, instead of the Luby sequence.
In other words, this is SGMPCS with a singleton elite solution.
We denote the first method Texture-Luby and the second method Texture-Geom+Guided.
These two methods were run on the same hardware with the same 
time limit and number of random runs as our method.
Finally, we report results for our approach without the greedy initialization 
heuristic (Algorithm~\ref{greedy}) in order to evaluate its importance.


We used the benchmarks generated by Caumond et al. in \cite{DBLP:journals/cor/CaumondLT08}
by adding maximal time lag constraints to the Lawrence JSP instances of the OR-library\footnote{\url{http://people.brunel.ac.uk/~mastjjb/jeb/info.html}}.
Given a job shop instance \texttt{N}, and two parameters $x$ and $y$,
a new instance \texttt{N}\_$x$\_$y$ is produced.
 For each job all 
maximal time lags are given 
the value $ym$, where $m$ is the average processing time
over tasks of this job. The first parameter $x$ corresponds to minimal time lags and will always be 
0 in this paper.


\begin{table}[htbp]
\footnotesize
\caption{ TL-JSP - Comparison with related work (Time \& Upper bound). \label{tab:TLJSPS} }
\begin{center}
\begin{tabular}{|l|| r  r | | r  r | | r  r | | }
\hline \hline
\multirow{2}{*}{Instance} & \multicolumn{ 2 }{c||}{ [AHL] } & \multicolumn{ 2 }{c||}{ [CLT] } & \multicolumn{ 2 }{c||}{ Model~\ref{mod:jtl} } \\ 
& \multicolumn{ 1 }{c}{ time (s) } & \multicolumn{ 1 }{c||}{ $\mkp$ } & \multicolumn{ 1 }{c}{ time (s) } & \multicolumn{ 1 }{c||}{ $\mkp$ } & \multicolumn{ 1 }{c}{ time (s) } & \multicolumn{ 1 }{c||}{ $\mkp$ } \\ 
 \hline 
\texttt{la06\_0\_10} ~&~       707.00 ~&~          927 ~&~         0.00 ~&~ \textbf{926} ~&~         0.03 ~&~ \textbf{926} ~\\
\texttt{la06\_0\_1}  ~&~       524.00 ~&~         1391 ~&~      1839.00 ~&~         1086 ~&~        70.60 ~&~ \textbf{926} ~\\
\texttt{la07\_0\_10} ~&~       518.00 ~&~         1123 ~&~        25.00 ~&~ \textbf{890} ~&~      3600.00 ~&~ \textbf{890} ~\\
\texttt{la07\_0\_1}  ~&~       754.00 ~&~         1065 ~&~      1914.00 ~&~         1032 ~&~      3600.00 ~&~ \textbf{896} ~\\
\texttt{la08\_0\_10} ~&~       260.00 ~&~ \textbf{863} ~&~         2.00 ~&~ \textbf{863} ~&~         0.07 ~&~ \textbf{863} ~\\
\texttt{la08\_0\_1}  ~&~       587.00 ~&~         1052 ~&~      1833.00 ~&~         1048 ~&~       615.80 ~&~ \textbf{892} ~\\
\hline 
 average ~&~       558.33 ~&~         1070 ~&~       935.50 ~&~          974 ~&~      1314.41 ~&~          898 ~\\
\hline 
 $PRD$ ~&~ ~&~       18.88 ~&~ ~&~       8.32 ~&~ ~&~       0.00 ~\\
\hline\hline
\end{tabular}
\end{center}
\end{table}

Due to space limitations, we present most of our results in terms of each solver's average percentage relative deviation (PRD)
 given by the following formula:
$
PRD = (( C_{Alg} - C_{Ref})/C_{Ref}) * 100 
$, where $C_{Alg}$ is the best makespan found by the algorithm and $C_{Ref}$ is the best upper bound among all 
considered algorithms\footnote{To the best of our knowledge, these are the best known upper bounds.}. 
In Table~\ref{tab:TLJSPS}, we first report a comparison with the genetic algorithm described in
\cite{DBLP:journals/cor/CaumondLT08}, denoted [CLT]
and the adhoc Branch \& Bound algorithm introduced in
\cite{Artigues2011220}, denoted [AHL].
We used only instances for which results were reported in both
papers, and where the time lags were strictly positive, hence
the relatively small data set.
Despite that, and despite the difference in hardware and time limit, 
it is quite clear that our approach
outperforms both the complete and heuristic methods on these benchmarks.



\begin{table}[htbp]
\footnotesize
\centering
\caption{TL-JSP - Comparison with Ilog Scheduler (Proofs of optimality \& Upper bound PRD). \label{tab:TLJSPIS} }
\begin{tabular}{|l|| r  r | r  r | | r r | r  r | | }
  \hline \hline
  \multirow{3}{*}{Instance Sets} & 
  \multicolumn{ 4 }{c||}{Texture} & 
  \multicolumn{ 4 }{c||}{Model~\ref{mod:jtl}} \\ 
  & 
  \multicolumn{ 2 }{c|}{Luby} & 
  \multicolumn{ 2 }{c||}{Geom+Guided} & 
  \multicolumn{ 2 }{c|}{ no init.  } & 
  \multicolumn{ 2 }{c||}{ init. heuristic } \\
  & 
  \multicolumn{ 1 }{c}{ Opt. } & 
  \multicolumn{ 1 }{c|}{ $PRD$ } & 
  \multicolumn{ 1 }{c}{ Opt. } & 
  \multicolumn{ 1 }{c||}{ $PRD$ } & 
  \multicolumn{ 1 }{c}{ Opt. } & 
  \multicolumn{ 1 }{c|}{ $PRD$ } & 
  \multicolumn{ 1 }{c}{ Opt. } & 
  \multicolumn{ 1 }{c||}{ $PRD$ } \\ 
  \hline
  \texttt{la[1,40]\_0\_0} ~&~         0.12 ~&~       25.37 ~&~         0.12 ~&~       16.15 ~&~ \textbf{0.37} ~&~       10.42 ~&~         0.35 ~&~ \textbf{0.06} ~\\
  \texttt{la[1,40]\_0\_0.25} ~&~         0.20 ~&~       22.98 ~&~         0.25 ~&~      12.01 ~&~         0.37 ~&~       3.46 ~&~ \textbf{0.40} ~&~ \textbf{0.00} ~\\
  \texttt{la[1,40]\_0\_0.5}  ~&~         0.22 ~&~       19.47 ~&~         0.25 ~&~       5.17 ~&~         0.37 ~&~       2.62 ~&~ \textbf{0.42} ~&~ \textbf{0.00} ~\\
  \texttt{la[1,40]\_0\_1} ~&~         0.35 ~&~       15.76 ~&~         0.42 ~&~       1.18 ~&~         0.40 ~&~       17.43 ~&~ \textbf{0.45} ~&~ \textbf{0.47} ~\\
  \texttt{la[1,40]\_0\_2} &~         0.67 ~&~       7.35 ~&~ \textbf{0.75} ~&~ \textbf{0.13} ~&~         0.67 ~&~       74.16 ~&~         0.70 ~&~       0.37 ~\\
  \texttt{la[1,40]\_0\_3} ~&~         0.75 ~&~       3.47 ~&~ \textbf{0.92} ~&~ \textbf{0.00} ~&~         0.75 ~&~       95.91 ~&~         0.77 ~&~       0.29 ~\\
  \texttt{la[1,40]\_0\_10} ~&~         0.95 ~&~       0.10 ~&~ \textbf{0.97} ~&~ \textbf{0.00} ~&~         0.92 ~&~       0.04 ~&~         0.92 ~&~       0.05 ~\\
  \hline
 \hline 
\end{tabular}
\end{table}

Next, in Table~\ref{tab:TLJSPIS}, we report results on all modified Lawrence instances for both
Ilog Scheduler models, and the two version of Model~\ref{mod:jtl},
with and without the greedy initialization heuristic.
 Since there are 280 instances in total, 
the results are
aggregated by the level of tightness of the time lag constraints.
For each set, we give the ratio of instances that were solved
to optimality in at least one of the five runs in the first column, as well as the
mean PRD in the second column.

First, we notice the great impact of the new initialization
heuristic on our method. Without it, the Ilog Scheduler model
was more efficient for instances with $y=1$, and the overall results
are extremely poor for larger values of $y$. However, the mean results 
are deceptive. Without initialization, Model~\ref{mod:jtl}
can be very efficient, although in a few cases no solution at all can be found.
Indeed, relaxing the makespan does not necessarily makes the problem
easy for this model. The weight of these bad cases in the mean value 
can be important, hence the poor PRD.
On the other hand, we can see that the Ilog Scheduler model is more robust
to this phenomenon: a non-trivial upper bound is found in every case.
It is therefore likely that the impact of the initialization heuristic
will not be as important on the Ilog model as on Model~\ref{mod:jtl}.

We also notice that solution guidance and 
geometric restarts greatly improve Ilog Scheduler's performance.
Interestingly, we observe that our approach is best when the time lag 
constraints are tight. On the other hand, Scheduler is slightly
more efficient on instances with loose time lag constraints 
and in particular proves optimality
more often on these instances. However, whereas our method always finds 
near-optimal solutions (the worst mean PRD is 0.47 for instances with $y=1$),
both scheduler models find relatively poor upper bounds for small values
of $y$.

\subsection{Job Shop Scheduling Problem with no wait constraints}

For the no-wait job shop problem, the best methods are a tabu search method
by Schuster (TS~\cite{Schuster06})
and a hybrid constructive/tabu search algorithm introduced by Bo\.zejko and Makuchowski in 2009 
(HTS~\cite{Bozejko20091502}). 
We also report the results of a  Branch \& Bound procedure introduced by
Mascis and Pacciarelli \cite{Mascis02}. This algorithm was run on a 
Pentium II 350 MHz.


\begin{table}[htbp] 
\footnotesize
\caption{ NW-JSP - Comparison with related work (Upper bound PRD). \label{tab:TLJSPRW} }
\begin{center}
\begin{tabular}{|l|| r | | r | r r | r r | r r | }
\hline \hline
\multirow{3}{*}{Instance} 
& \multicolumn{ 1 }{c|}{Mascis et al.} & \multicolumn{ 1 }{c|}{Schuster} & \multicolumn{ 2 }{c|}{Bo\.zejko et al.}  & \multicolumn{ 2 }{c|}{ Model~\ref{mod:now1} }  & \multicolumn{ 2 }{c|}{ Model~\ref{mod:now2} }\\ 
& \multicolumn{ 1 }{c|}{ B\&B } & \multicolumn{ 1 }{c|}{ TS } & \multicolumn{ 1 }{c}{HTS} & \multicolumn{ 1 }{c|}{HTS+} & \multicolumn{ 1 }{c}{ $tdom$+$bw$ } & \multicolumn{ 1 }{c|}{ $tdom/tw$ } & \multicolumn{ 1 }{c}{ $tdom$+$bw$ } & \multicolumn{ 1 }{c|}{ $tdom/tw$ }\\ 

 \hline 
\texttt{la[1-10]} &~        {\bf 0.00} &~        4.43 &~        1.77 &~        {\it 1.77} &~        {\bf 0.00} &~        {\bf 0.00} &~        {\bf 0.00} &~        {\bf 0.00} \\
\texttt{la[11-20]} &~        31.66 &~        7.93 &~        3.49 &~        {\it 0.95} &~        0.14 &~        0.10 &~        {\bf 0.00} &~        0.31 \\
\texttt{la[21-30]} &~        61.09 &~        10.43 &~        7.25 &~        {\it 0.08} &~        1.16 &~        0.57 &~        {\bf 0.25} &~        0.84 \\
\texttt{la[31-40]} &~        73.73 &~        10.95 &~        8.33 &~        {\it 0.15} &~        4.42 &~        1.77 &~        2.68 &~        {\bf 1.36} \\
\texttt{abz[5-9]} &~        47.04 &~        9.01 &~        5.95 &~        {\it 0.78} &~        2.47 &~        1.14 &~        {\bf 1.13} &~        1.20 \\
\texttt{orb[1-10]} &~        {\bf 0.00} &~        2.42 &~        0.77 &~        {\it 0.77} &~        {\bf 0.00} &~        {\bf 0.00} &~        {\bf 0.00} &~        {\bf 0.00} \\
\texttt{swv[1-5]} &~        60.85 &~        3.94 &~        3.67 &~        {\it 0.00} &~        2.54 &~        0.77 &~        {\bf 0.00} &~        0.43 \\
\texttt{swv[6-10]} &~        57.82 &~        4.99 &~        4.19 &~        {\it 0.00} &~        4.78 &~        1.71 &~        {\bf 0.44} &~        1.00 \\
\texttt{swv[11-15]} &~        70.98 &~        {\bf 0.68} &~        2.48 &~        {\it 0.60} &~        19.50 &~        6.53 &~        17.54 &~        5.18 \\
\texttt{swv[16-20]} &~        76.81 &~        5.71 &~        3.98 &~        {\it 0.00} &~        10.92 &~        68.94 &~        4.47 &~        {\bf 3.17} \\
\texttt{yn[1-4]} &~        72.74 &~        12.40 &~        8.85 &~        {\it 0.32} &~        5.60 &~        5.75 &~        {\bf 2.37} &~        2.88 \\
\hline
overall &~        44.72 &~        6.51 &~        4.36 &~        {\it 0.52} &~        3.53 &~        5.50 &~        1.97 &~        {\bf 1.13} \\
\hline\hline
\end{tabular}
\end{center}
\end{table}

For the no-wait class 
we used the same data sets as Schuster~\cite{Schuster06}
and Bo\.zejko et al.~\cite{Bozejko20091502} where null time lags are added to instances of the OR-library.
We report the best results of each paper in terms of average PRD. 
It should be noted that for HTS,
the authors reported two sets of results. The 
former were run with a time limit based on the
runtimes reported in \cite{Schuster06} and varying from
0.25 seconds for the easiest instances to $2360$ seconds
for the hardest.
The latter (in italic font, and referred to as HTS+ in Table~\ref{tab:TLJSPRW})
 were run ``without limit of computation time''. We use bold face to mark the best result
amongst methods that had time limits, i.e. excluding HTS+.
We ran two variable ordering heuristics for our method. First, the 
heuristics used for ET-JSP and TL-JSP, where the Boolean variable 
minimizing the value of 
$
(max(\task{i}) + max(\task{j}) - min(\task{i}) - min(\task{j}) + 2) / (w(\task{i}) + w(\task{j}))
$ is chosen first, denoted $tdom/tw$.
Second, we used another heuristic, denoted $tdom$+$bw$ that selects the next Boolean variable
to branch on solely according to the tasks' domain sizes 
$(max(\task{i}) + max(\task{j}) - min(\task{i}) - min(\task{j}) + 2)$, and break ties
with the Boolean variable's own weight $w(\disjunct{ij})$.

\begin{table}[htbp]
\footnotesize
\caption{ NW-JSP - New best upper bounds and optimality proofs. \label{tab:TLJSPNB} }
\begin{center}
\begin{tabular}{|l|| r | | r | r r | r r | r r | }
\hline \hline
\multirow{3}{*}{Instance} 
& \multicolumn{ 1 }{c|}{} & \multicolumn{ 1 }{c|}{Schuster} & \multicolumn{ 2 }{c|}{Bo\.zejko}   & \multicolumn{ 2 }{c|}{ Model~\ref{mod:now1} }  & \multicolumn{ 2 }{c|}{ Model~\ref{mod:now2} }\\ 
& \multicolumn{ 1 }{c|}{ BKS } & \multicolumn{ 1 }{c|}{ TS } & \multicolumn{ 1 }{c}{HTS} & \multicolumn{ 1 }{c|}{HTS+} & \multicolumn{ 1 }{c}{ $tdom$+$bw$ } & \multicolumn{ 1 }{c|}{ $tdom/tw$ } & \multicolumn{ 1 }{c}{ $tdom$+$bw$ } & \multicolumn{ 1 }{c|}{ $tdom/tw$ }\\ 
 \hline 
la11\_0\_0 ~&~         2821 ~&~         1737 ~&~         1704 ~&~         1621 ~&~         1622 ~&~ \textbf{1619} ~&~ {\textbf{1619*}} ~&~         1621 ~\\
la13\_0\_0 ~&~         2650 ~&~         1701 ~&~         1696 ~&~ \textbf{1580} ~&~         1582 ~&~         1590 ~&~ {\textbf{1580*}} ~&~ \textbf{1580} ~\\
la14\_0\_0 ~&~         2662 ~&~         1771 ~&~         1722 ~&~         1610 ~&~ \textbf{1578} ~&~ \textbf{1578} ~&~ {\textbf{1578*}} ~&~         1612 ~\\
la15\_0\_0 ~&~         2765 ~&~         1808 ~&~         1747 ~&~         1686 ~&~         1692 ~&~         1679 ~&~ {\textbf{1671*}} ~&~         1691 ~\\
la26\_0\_0 ~&~         4268 ~&~         2664 ~&~         2738 ~&~         2506 ~&~         2624 ~&~         2511 ~&~ \textbf{2488} ~&~         2540 ~\\
la28\_0\_0 ~&~         4478 ~&~         2886 ~&~         2741 ~&~         2552 ~&~         2640 ~&~         2605 ~&~ \textbf{2546} ~&~         2569 ~\\
la30\_0\_0 ~&~         4097 ~&~         2939 ~&~         2791 ~&~ \textbf{2452} ~&~ \textbf{2452} ~&~ \textbf{2452} ~&~ {\textbf{2452*}} ~&~         2508 ~\\
la34\_0\_0 ~&~         6380 ~&~         3957 ~&~         3936 ~&~         3659 ~&~         3914 ~&~         3693 ~&~         3817 ~&~ \textbf{3657} ~\\
la39\_0\_0 ~&~         4295 ~&~         2804 ~&~         2725 ~&~         2687 ~&~ \textbf{2660} ~&~ \textbf{2660} ~&~ {\textbf{2660*}} ~&~ \textbf{2660} ~\\
swv01      ~&~         3824 ~&~         2396 ~&~         2424 ~&~ \textbf{2318} ~&~         2344 ~&~         2343 ~&~ {\textbf{2318*}} ~&~         2333 ~\\
swv02      ~&~         3800 ~&~         2492 ~&~         2484 ~&~ \textbf{2417} ~&~         2440 ~&~         2418 ~&~ {\textbf{2417*}} ~&~ \textbf{2417} ~\\
swv05      ~&~         3836 ~&~         2482 ~&~         2489 ~&~ \textbf{2333} ~&~         2433 ~&~ \textbf{2333} ~&~ {\textbf{2333*}} ~&~ \textbf{2333} ~\\
yn2        ~&~         4025 ~&~         2705 ~&~         2647 ~&~         2370 ~&~         2486 ~&~         2603 ~&~         2427 ~&~ \textbf{2353} ~\\
yn4        ~&~         4109 ~&~         2705 ~&~         2630 ~&~         2513 ~&~         2532 ~&~         2573 ~&~ \textbf{2499} ~&~         2582 ~\\
\hline\hline
\end{tabular}
\end{center}
\end{table}


In Table~\ref{tab:TLJSPNB} we report the results on no-wait instances for which we obtained new upper bounds (5 instances) 
or new proofs of optimality (9 instances), thanks to the model introduced here.

\section{Conclusions}

We have shown that the simple constraint programming approach 
introduced in \cite{grimes09:clo} can be successfully 
adapted to handle the
objective of minimizing the sum of earliness/tardiness costs. These
problems have traditionally proven troublesome for CP approaches
because of the weak propagation of the sum objective
\cite{Danna03lnsTechSchedetjsp}.

Then we introduced a new heuristic to find good initial solutions
for job shop problems with maximal time lag constraints.
The resulting method greatly improves over state of the art
algorithms for this problem. However, as opposed to the 
other aspects of the method (adaptive variable heuristic,
solution guided branching, restarts with nogood storage)
this new initialization heuristic is dedicated to job shop
problems with time lag constraints.

Finally, we showed that domain-specific information can also be used to improve our model for no-wait job shop scheduling 
problems, allowing us to provide several improved upper bounds and 
prove optimality in many cases.

\bibliographystyle{plain}
\bibliography{ospbibdgnew}

\begin{thebibliography}{10}

\bibitem{Artigues2011220}
C.~Artigues, M-J. Huguet, and P.~Lopez.
\newblock {Generalized Disjunctive Constraint Propagation for Solving the Job
  Shop Problem with Time Lags}.
\newblock {\em EAAI}, 24(2):220 -- 231, 2011.

\bibitem{Baptiste08SchedJIT}
P.~Baptiste, M.~Flamini, and F.~Sourd.
\newblock {Lagrangian Bounds for Just-in-Time Job-shop Scheduling}.
\newblock {\em Computers {\&} OR}, 35(3):906--915, 2008.

\bibitem{Becksgmp}
J.~C. Beck.
\newblock Solution-{G}uided {M}ulti-{P}oint {C}onstructive {S}earch for {J}ob
  {S}hop {S}cheduling.
\newblock {\em JAIR}, 29:49--77, 2007.

\bibitem{Beck03Schedetjsp}
J.~C. Beck and P.~Refalo.
\newblock {A Hybrid Approach to Scheduling with Earliness and Tardiness Costs}.
\newblock {\em Annals OR}, 118(1-4):49--71, 2003.

\bibitem{Boussemart-etal:2004b}
F.~Boussemart, F.~Hemery, C.~Lecoutre, and L.~Sais.
\newblock Boosting {S}ystematic {S}earch by {W}eighting {C}onstraints.
\newblock In {\em ECAI}, pages 482--486, 2004.

\bibitem{Bozejko20091502}
W.~Bozejko and M.~Makuchowski.
\newblock {A Fast Hybrid Tabu Search Algorithm for the No-wait Job Shop
  Problem}.
\newblock {\em Computers \& Industrial Engineering}, 56(4):1502--1509, 2009.

\bibitem{DBLP:journals/cor/CaumondLT08}
A.~Caumond, P.~Lacomme, and N.~Tchernev.
\newblock {A Memetic Algorithm for the Job-shop with Time-lags}.
\newblock {\em Computers {\&} OR}, 35(7):2331--2356, 2008.

\bibitem{Danna03lnsTechSchedetjsp}
E.~Danna and L.~Perron.
\newblock Structured vs. unstructured large neighborhood search: A case study
  on job-shop scheduling problems with earliness and tardiness costs.
\newblock Technical report, ILOG, 2003.

\bibitem{Danna03rinsSchedetjsp}
E.~Danna, E.~Rothberg, and C.~Le Pape.
\newblock {Integrating Mixed Integer Programming and Local Search: A Case Study
  on Job-Shop Scheduling Problems}.
\newblock In {\em CPAIOR}, 2003.

\bibitem{lazy-fd}
T.~Feydy and P.~J. Stuckey.
\newblock Lazy {C}lause {G}eneration {R}eengineered.
\newblock In {\em CP}, pages 352--366, 2009.

\bibitem{grimes08:stu}
D.~Grimes.
\newblock {A Study of Adaptive Restarting Strategies for Solving Constraint
  Satisfaction Problems}.
\newblock In {\em AICS}, 2008.

\bibitem{grimes10:job}
D.~Grimes and E.~Hebrard.
\newblock {J}ob {S}hop {S}cheduling with {S}etup {T}imes and {M}aximal
  {T}ime-{L}ags: {A} {S}imple {C}onstraint {P}rogramming {A}pproach.
\newblock In {\em CPAIOR}, pages 147--161, 2010.

\bibitem{grimes09:clo}
D.~Grimes, E.~Hebrard, and A.~Malapert.
\newblock Closing the {O}pen {S}hop: {C}ontradicting {C}onventional {W}isdom.
\newblock In {\em CP'09}, pages 400--408, 2009.

\bibitem{Kelbel10etjspSched}
J.~Kelbel and Z.~Hanz{\' a}lek.
\newblock {Solving production scheduling with earliness/tardiness penalties by
  constraint programming}.
\newblock {\em J. Intell. Manuf.}, 2010.

\bibitem{LecoutreSTV07}
C.~Lecoutre, L.~Sais, S.~Tabary, and V.~Vidal.
\newblock Nogood {R}ecording from {R}estarts.
\newblock In {\em IJCAI}, pages 131--136, 2007.

\bibitem{LubySZ93}
M.~Luby, A.~Sinclair, and D.~Zuckerman.
\newblock Optimal {S}peedup of {L}as {V}egas {A}lgorithms.
\newblock In {\em ISTCS}, pages 128--133, 1993.

\bibitem{Mascis02}
A.~Mascis and D.~Pacciarelli.
\newblock {Job-shop Scheduling with Blocking and No-wait Constraints}.
\newblock {\em EJOR}, 143(3):498--517, 2002.

\bibitem{Morton93DynJspSched}
T.~E. Morton and D.~W. Pentico.
\newblock {\em Heuristic Scheduling Systems}.
\newblock John Wiley and Sons, 1993.

\bibitem{rajendran94:ano}
C.~Rajendran.
\newblock {A No-Wait Flowshop Scheduling Heuristic to Minimize Makespan}.
\newblock {\em The Journal of the Operational Research Society},
  45(4):472--478, 1994.

\bibitem{Schuster06}
C.~J. Schuster.
\newblock {No-wait Job Shop Scheduling: Tabu Search and Complexity of
  Problems}.
\newblock {\em Math Meth Oper Res}, 63:473--491, 2006.

\bibitem{schutt09:why}
A.~Schutt, T.~Feydy, P.~J. Stuckey, and M.~Wallace.
\newblock {Why Cumulative Decomposition Is Not as Bad as It Sounds}.
\newblock In {\em CP'09}, pages 746--761, 2009.

\bibitem{tamura06:com}
N.~Tamura, A.~Taga, S.~Kitagawa, and M.~Banbara.
\newblock {Compiling Finite Linear CSP into SAT}.
\newblock In {\em CP}, pages 590--603, 2006.

\bibitem{Vazquez00etjspSched}
M.~V{\'a}zquez and L.~D. Whitley.
\newblock A comparison of genetic algorithms for the dynamic job shop
  scheduling problem.
\newblock In {\em GECCO}, pages 1011--, 2000.

\bibitem{Vilim2008Unary}
P.~Vil\'{i}m.
\newblock {Filtering Algorithms for the Unary Resource Constraint}.
\newblock {\em Archives of Control Sciences}, 18(2), 2008.

\bibitem{Walsh99}
T.~Walsh.
\newblock Search in a {S}mall {W}orld.
\newblock In {\em IJCAI}, pages 1172--1177, 1999.

\bibitem{Watson99JspGeneratorSched}
J-P. Watson, L.~Barbulescu, A.~E. Howe, and L.~D. Whitley.
\newblock Algorithm performance and problem structure for flow-shop scheduling.
\newblock In {\em AAAI}, pages 688--695, 1999.

\bibitem{wismer72:sol}
D.~A. Wismer.
\newblock {Solution of the Flowshop-Scheduling Problem with No Intermediate
  Queues}.
\newblock {\em Operations Research}, 20(3):689--697, 1972.

\end{thebibliography}

\end{document}